\documentclass[10pt,twocolumn,letterpaper]{article}

\usepackage{cvpr}
\usepackage{times}
\usepackage{epsfig}
\usepackage{graphicx}
\usepackage{amsmath}
\usepackage{amssymb}

\usepackage{multirow}
\usepackage{cmath}
\usepackage{subfigure}

\usepackage[breaklinks=true,bookmarks=false]{hyperref}

\cvprfinalcopy 

\def\cvprPaperID{4919} 

\ifcvprfinal\fi
\begin{document}

\title{Domain-aware Visual Bias Eliminating for Generalized Zero-Shot Learning}

\author{Shaobo Min\textsuperscript{1}, Hantao Yao\textsuperscript{2}, Hongtao Xie\textsuperscript{1}\thanks{Corresponding author.}, Chaoqun Wang\textsuperscript{1}, Zheng-Jun Zha\textsuperscript{1}, and Yongdong Zhang\textsuperscript{1}\\
\textsuperscript{1}University of Science and Technology of China\\
\textsuperscript{2}National Laboratory of Pattern Recognition, Institute of Automation, Chinese Academy of Sciences\\
{\tt\small \{mbobo,cq14\}@mail.ustc.edu.cn, hantao.yao@nlpr.ia.ac.cn, \{htxie,zhazj,zhyd73\}@ustc.edu.cn}
}

\maketitle

\begin{abstract}
Generalized zero-shot learning aims to recognize images from seen and unseen domains. 
Recent methods focus on learning a unified semantic-aligned visual representation to transfer knowledge between two domains, while ignoring the effect of semantic-free visual representation in alleviating the biased recognition problem.
In this paper, we propose a novel Domain-aware Visual Bias Eliminating (DVBE) network that constructs two complementary visual representations,~\emph{i.e.,} semantic-free and semantic-aligned, to treat seen and unseen domains separately.
Specifically, we explore cross-attentive second-order visual statistics to compact the semantic-free representation, and design an adaptive margin Softmax to maximize inter-class divergences.
Thus, the semantic-free representation becomes discriminative enough to not only predict seen class accurately but also filter out unseen images,~\emph{i.e.,} domain detection, based on the predicted class entropy.
For unseen images, we automatically search an optimal semantic-visual alignment architecture, rather than manual designs, to predict unseen classes.
With accurate domain detection, the biased recognition problem towards the seen domain is significantly reduced.
Experiments on five benchmarks for classification and segmentation show that DVBE outperforms existing methods by averaged $5.7\%$ improvement.
\end{abstract}


\section{Introduction}
Deep learning has made great progress in many vision tasks, such as image classification \cite{gu2018asynchronous,liu2016hierarchical,8907499,yao2017autobd,yao2016coarse,zheng2018fast}, object detection \cite{law2018cornernet,ren2015faster}, and semantic segmentation \cite{chen2017deeplab,zhang2017scale,zhang2019perspective}.
However, these remarkable technologies depend heavily on large-scale data of various classes, and it is infeasible to annotate them all.
As a consequence, generalized zero-shot learning (GZSL) has attracted increasing attention, which can recognize images from either seen or unseen classes.
\begin{figure}
	\begin{center}
		\includegraphics[width=1\linewidth]{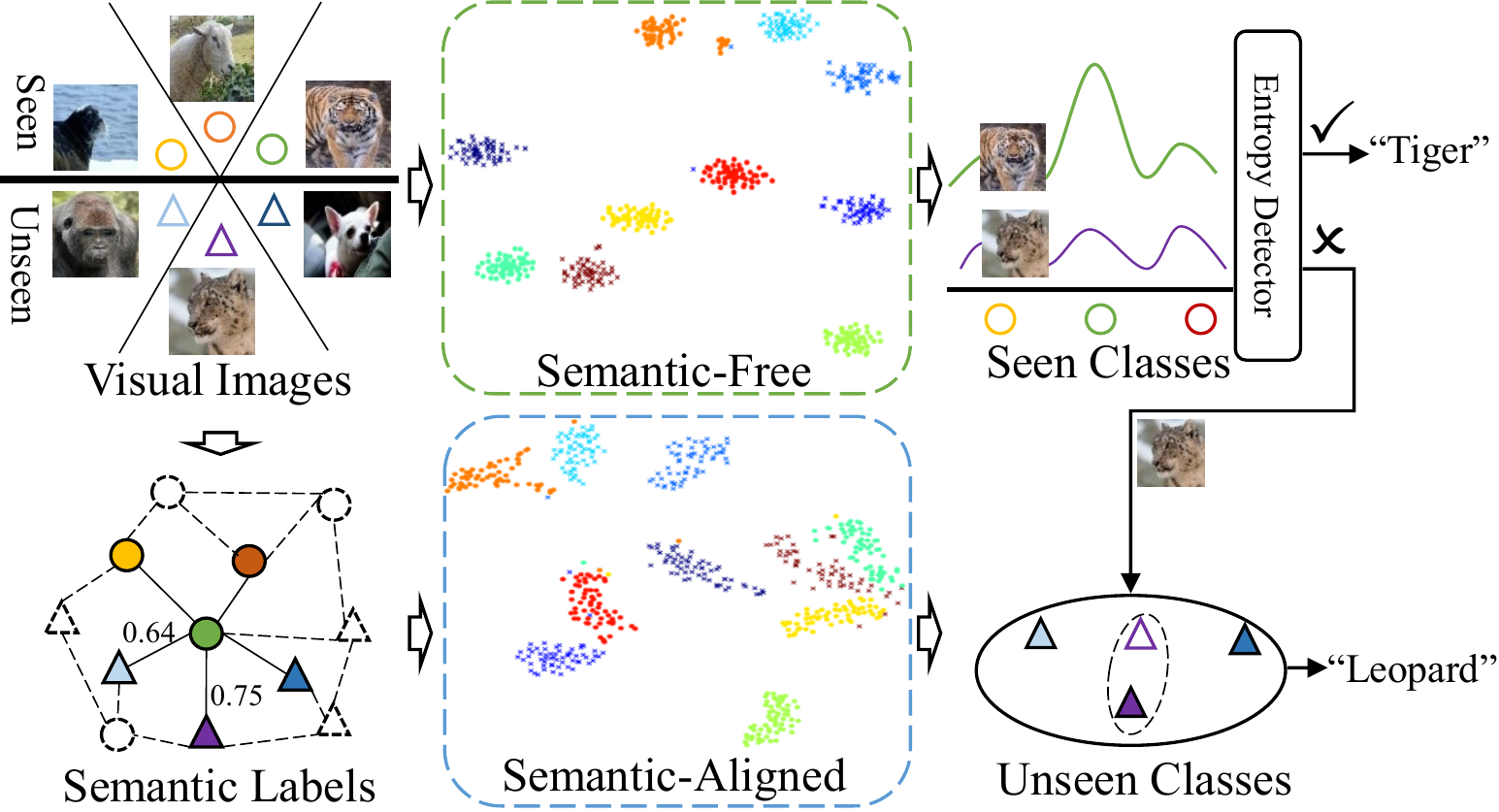}
	\end{center}
	\caption{The inference framework of DVBE. Due to the weak discrimination of semantic-aligned visual representation, DVBE constructs an extra semantic-free visual representation to improve seen class prediction and filter out unseen images correctly.
	}
	\label{fig:moti}
\vspace{-0.5cm} 
\end{figure}

A general paradigm of GZSL is to align the visual images and semantic labels, \emph{e.g.,} category attributes \cite{Farhadi2009}, in a joint semantic space, where the recognition becomes a nearest neighbor searching problem \cite{Xian2016,xian2017zero,zhu2019generalized}.
In the joint space, semantic embeddings and visual representations are treated as class anchors and input queries, respectively \cite{Annadani2018,ding2019marginalized}. 
However, the provided semantic labels are usually not discriminative~\cite{ding2019marginalized,paul2019semantically,zhu2019generalized}, some of which are much similar,~\emph{e.g.,} $0.75$ cosine similarity between ``Tiger" and ``Leopard" in Figure~\ref{fig:moti}. 
Thus, after semantic-visual alignment, the inter-class divergence between visual representations is inevitably weakened, which further results in the biased recognition problem \cite{ding2019marginalized,elhoseiny2019creativity,tong2019hierarchical},~\emph{i.e.,} unseen images tend to be recognized as seen classes.

To address the above issue, recent methods~\cite{Annadani2018,Chen2018,Li_2019_ICCV,tong2019hierarchical,zhu2019generalized,zhu2019semantic} focus on enlarging the visual representation discrimination.
For example, SP-AEN~\cite{Chen2018} preserves the discriminative visual topology by reconstructing visual representations back to images.
DLFZRL~\cite{tong2019hierarchical} factorizes an image representation into several latent components to improve model generalization.
AREN~\cite{xie2019attentive} and VSE~\cite{zhu2019generalized} explore part-based embeddings to capture detailed visual clues.
Although being effective, the visual representations of these methods are ultimately aligned with semantic labels for knowledge transfer.
Therefore, they still suffer from degraded visual discrimination by semantic-visual alignment. 
Different from the previous methods, we consider learning an extra semantic-free visual representation in GZSL.
Supervised with only class labels, the highly discriminative semantic-free representation can not only improve seen class prediction but also filter out unseen images based on seen class prediction entropy,~\emph{i.e.,} domain detection, as shown in Figure~\ref{fig:moti}, though it cannot be transferred to the unseen domain.
Once the domain detection is accurate, the filtered unseen images can be recognized in the domain-aware searching space, thereby alleviating the biased recognition problem.

\begin{figure*}
	\begin{center}
		\includegraphics[width=0.9\linewidth]{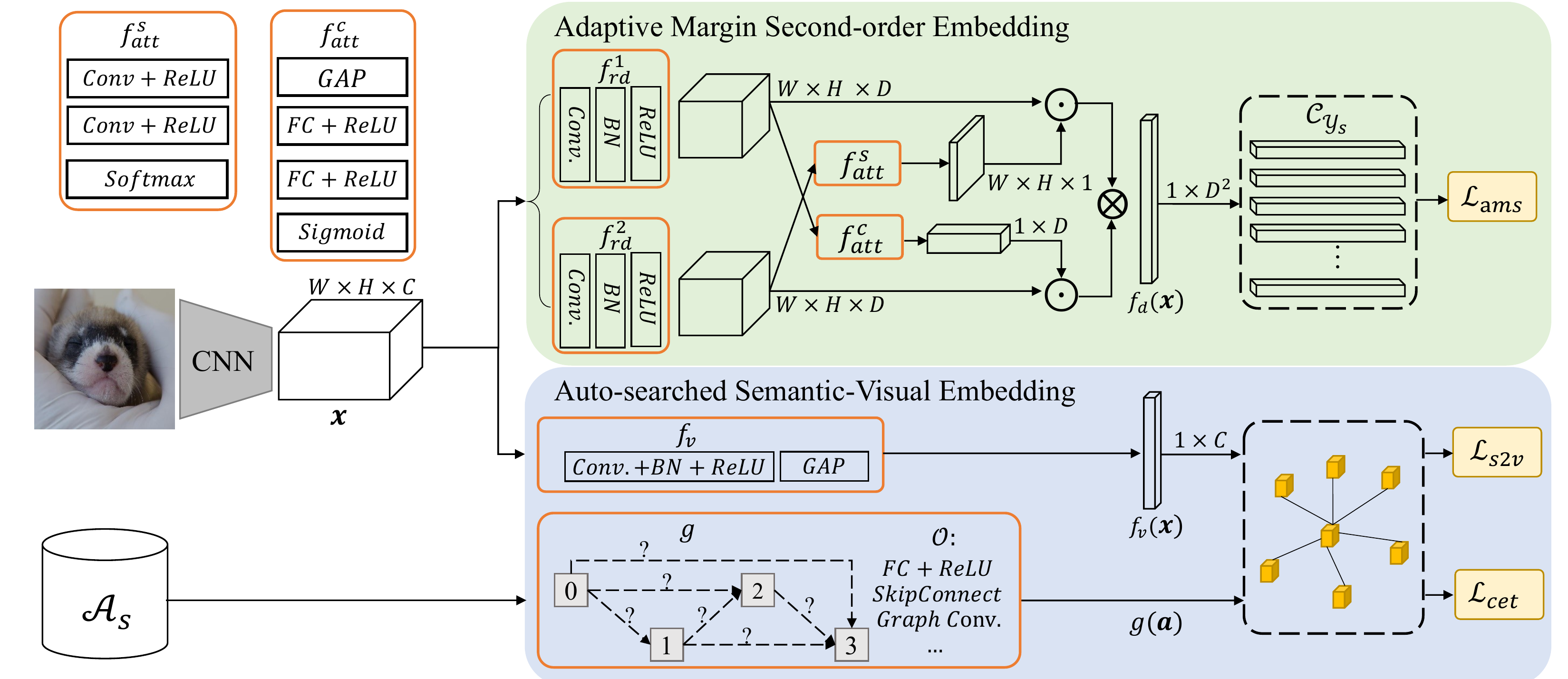}
	\end{center}
	\caption{The training framework of DVBE with detailed implementation. $GAP$ is global average pooling, and DVBE is trained with only seen domain data.}
	\label{fig:method}
\vspace{-0.4cm} 
\end{figure*}

Based on the above discussion, we propose a novel Domain-aware Visual Bias Eliminating (DVBE) network to construct two complementary visual representations, which are semantic-free and semantic-aligned, respectively.
Explicitly, DVBE focuses on two main problems: a) how to boost the discrimination of semantic-free representation to improve domain detection accuracy; and b) how to design the semantic-visual alignment architecture for robust knowledge transfer.
To this end, DVBE consists of two sub-modules,~\emph{i.e.,} adaptive margin second-order embedding (AMSE), and auto-searched semantic-visual embedding (autoS2V), as shown in Figure~\ref{fig:method}.

With only seen domain training data, AMSE targets to learn a semantic-free representation that is discriminative enough to classify seen images and filter out unseen images correctly.
To this end, AMSE explores second-order visual statistics to capture subtle inter-class differences.
To reduce the trivial information in high-dimensional second-order representations, a cross-attentive channel interaction is designed along both spatial and channel dimensions.
Besides, adaptive margin constraints are imposed between decision boundaries of different classes to maximize the inter-class discrepancy. 
Consequently, AMSE significantly reduces the entropy of predicted seen class scores for seen domain images.
Thus, unseen images of relatively high entropy can be correctly filtered out via an entropy-based detector. 
For the filtered unseen images, autoS2V is proposed to automatically search an optimal semantic-visual architecture via continuous architecture relaxation~\cite{liu2018darts}, which can produce robust semantic-aligned representation.
Compared with manual designs \cite{Annadani2018,zhu2019generalized}, autoS2V is more flexible in bridging the semantic-visual gap.

Experiments on object classification (averaged $4.3\%$ improvement on CUB \cite{WelinderEtal2010}, AWA2 \cite{Xian2018}, aPY \cite{Farhadi2009}, and SUN \cite{Patterson2012}) and semantic segmentation ($11\%$ improvement on Pascal VOC \cite{everingham2010pascal}) prove the effectiveness of DVBE.
Our contributions can be summarized by:
\begin{itemize}
\setlength{\itemsep}{0pt}
\setlength{\parsep}{0pt}
\setlength{\parskip}{0pt}

\item  We propose a novel Domain-aware Visual Bias Eliminating network that constructs two complementary visual representations,~\emph{i.e.,} semantic-free and semantic-aligned, to tackle respective seen and unseen domains for unbiased GZSL.

\item An adaptive margin second-order embedding is developed to generate highly discriminative semantic-free representation, which can improve seen class prediction and filter out unseen images accurately.

\item An auto-searched semantic-visual embedding is designed to automatically search an optimal semantic-visual architecture for robust knowledge transfer.

\end{itemize}

\section{Related Works}
A general paradigm of GZSL is to align image representations and extra semantic information in a joint embedding space \cite{Chen2018,Jiang_2018_ECCV,xu2018dual,zhu2019generalized}.
The commonly used semantic information is within manually annotated category attributes \cite{Farhadi2009,Morgado2017}, word2vec \cite{Frome2013,zhang2017learning}, and text descriptions \cite{lei2015predicting}.
Under this paradigm, the biased recognition problem \cite{ding2019marginalized,elhoseiny2019creativity,liu2019adaptive,tong2019hierarchical,zhang2012attribute},~\emph{i.e.,} unseen images tend to be recognized as seen classes, is a major obstacle.
One of the main reasons is that the provided semantic labels are usually not discriminative \cite{ding2019marginalized,li2018discriminative,zhu2019generalized}, which make the semantic-aligned visual representations from two domains hard to distinguish.
To enlarge the visual discrimination, previous methods \cite{min2019domain,Xian2018,zhu2019generalized} replace the embedding space, spanned by semantic labels, with the high dimensional visual space.
This prevents a few semantic embeddings of the seen domain from being nearest neighbors for most of the image samples \cite{Annadani2018}.
Later, the auto-encoder architecture is widely adopted \cite{kodirov2017semantic,tong2019hierarchical} to preserve discriminative latent information.
For example, SP-AEN \cite{Chen2018} reconstructs visual representations back to images to retain the visual topology.
PSR-ZSL \cite{Annadani2018} defines explicit semantic relations,~\emph{i.e.,} dissimilar, similar, and identical, during semantic reconstruction, and DLFZRL \cite{tong2019hierarchical} exploits auto-encoder to factorize an image into several latent features and fuse them into a generalized one. 
Furthermore, to improve the description ability of visual representations, part-based embeddings are used in \cite{xie2019attentive,zhu2019generalized,zhu2019semantic}.
By automatically localizing informative regions, they can capture a lot of important local clues to improve semantic-aligned representations, which achieve robust knowledge transfer. 

Recently, generative methods \cite{felix2018multi,kumar2018generalized,mishra2018generative,schonfeld2019generalized,xian2018feature,Xian_2019_CVPR} obtain impressive results by training powerful GANs \cite{Goodfellow2014} in the seen domain and directly generate massive unseen visual features from the semantic labels.
This allows them to train a fully supervised classifier for two domains, which is robust to the biased recognition problem.
In this paper, our focus is embedding-based GZSL, which requires no extra synthetic training data and complex training processing.

Although being effective, a commonality of the above embedding-based methods is that their visual representations are finally aligned with semantic labels for knowledge transfer, which are inevitably affected by the weak semantic discrimination.
Differently, the proposed DVBE explores the totally semantic-free visual representation and find it can alleviate the biased recognition problem.
Notably, COSMO \cite{atzmon2019adaptive} is the most similar work to us, which is also based on domain detection.
However, COSMO focuses on designing an elaborate domain detector to improve domain detection.
Differently, DVBE emphasizes on learning a discriminative semantic-free representation via AMSE to improve domain detection, and adopts a simple entropy-based detector for end-to-end training.
The experiments show that our DVBE is superior to COSMO.

\section{Domain-aware Visual Bias Eliminating}
Generalized zero-shot learning aims to recognize images from seen and unseen domains. 
Formally, we define the seen domain data as $\mathcal{S}=\{\boldsymbol{x}_s,y_s,\boldsymbol{a}_s|\boldsymbol{x}_s\in\mathcal{X}_s,y_s\in \mathcal{Y}_s,\boldsymbol{a}_s\in\mathcal{A}_s\}$, where $\boldsymbol{x}_s$ is the backbone feature for an image, $y_s$ is the class label, and $\boldsymbol{a}_s$ denotes the semantic label, such as class attributes or text descriptions.
Similarly, the unseen domain data is defined as $\mathcal{U}=\{\boldsymbol{x}_u,y_u,\boldsymbol{a}_u|\boldsymbol{x}_u\in\mathcal{X}_u,y_u\in \mathcal{Y}_u,\boldsymbol{a}_u\in\mathcal{A}_u\}$, where $\mathcal{Y}_s\cap\mathcal{Y}_u=\phi$.
Trained with seen domain data $\mathcal{S}$, GZSL targets to recognize a sample from ether $\mathcal{X}_s$ or $\mathcal{X}_u$.

A general paradigm of GZSL is to learn a semantic-visual space by minimizing:
\begin{eqnarray}
\mathcal{L}_{s2v} = \sum_{\boldsymbol{x}\in\mathcal{X}_s}d(f_{v}(\boldsymbol{x}),g(\boldsymbol{a}_y)),
\label{eq:gel_com}
\end{eqnarray}
where $y$ is the class label for sample $\boldsymbol{x}$, and $\boldsymbol{a}_y$ is the corresponding semantic label for class $y$.
$f_{v}(\cdot)$ and $g(\cdot)$ are the visual and semantic embedding functions to project $\boldsymbol{x}$ and $\boldsymbol{a}_y$ into the joint embedding space. 
The metric function $d(\cdot,\cdot)$ measures the distance between $f_{v}(\boldsymbol{x})$ and $g(\boldsymbol{a}_y)$,~\emph{e.g.,} cosine distance \cite{Annadani2018}.
Based on an assumption that seen and unseen domains share the same semantic space, the semantic-visual alignment inferred from $\mathcal{X}_s$ and $\mathcal{A}_s$ can be directly transferred to $\mathcal{X}_u$ and $\mathcal{A}_u$.
Thus, the inference of GZSL becomes a nearest neighbor searching problem which can be defined by:
\begin{eqnarray}
\hat{y} = \arg\min_{y\in\mathcal{Y}_s\cup\mathcal{Y}_u}d(f_{v}(\boldsymbol{x}),g(\boldsymbol{a}_y)),
\label{eq:gel_inf}
\end{eqnarray}
where $\boldsymbol{x}\in\mathcal{X}_s\cup\mathcal{X}_u$, and $\hat{y}$ is the predicted class.

Although the semantic-visual alignment constructed by $f_{v}(\cdot)$ and $g(\cdot)$ can be adapted to $\mathcal{U}$, the visual discrimination of $f_{v}(\boldsymbol{x})$ is disturbed by the semantic labels.
For example, if two categories have much similar semantic labels, the visual representations of images belonging to these two categories are also hard to distinguish, due to semantic-visual alignment in Eq.~\eqref{eq:gel_com}.
Unfortunately, the provided semantic labels usually have a small inter-class difference between two domains \cite{zhu2019generalized}, which makes the semantic-aligned $f_{v}(\boldsymbol{x}_s)$ and $f_{v}(\boldsymbol{x}_u)$ hard to distinguish and $f_{v}(\boldsymbol{x}_u)$ tend to be matched with $\boldsymbol{a}_s$.

\subsection{Formulation}
To tackle the above issues, we propose a novel Domain-aware Visual Bias Eliminating (DVBE) network.
Besides the semantic-aligned visual representation $f_{v}(\boldsymbol{x})$ in Eq.~\eqref{eq:gel_inf}, we introduce an extra semantic-free visual representation $f_{d}(\boldsymbol{x})$ that is trained on $\{\boldsymbol{x}_s,y_s\}$. 
Although $f_{d}(\boldsymbol{x})$ cannot be transferred to predict unseen classes without semantic relations, it can filter out the unseen images based on seen class prediction entropy, which enables $f_{v}(\boldsymbol{x})$ to focus on inferring the unseen classes.
Based on the above discussion, we converts Eq.~\eqref{eq:gel_inf} to:
\begin{eqnarray}\label{eq:inf_all}
\hat{y}=\left\{
\begin{aligned}
&\arg\max_{y\in\mathcal{Y}_s}\mathcal{C}_y(f_{d}(\boldsymbol{x})) \qquad\quad if~ \mathcal{H}(\mathcal{C}(f_{d}(\boldsymbol{x}))\leqslant\tau, \\
&\arg\min_{y\in\mathcal{Y}_u} d(f_{v}(\boldsymbol{x}),g(\boldsymbol{a}(y)) \quad\quad\quad else,\qquad~
\end{aligned}
\right.
\end{eqnarray}
where $f_{d}(\boldsymbol{x})$ and $f_{v}(\boldsymbol{x})$ are semantic-free and semantic-aligned embedding functions, respectively.
$\mathcal{C}$ is a $|\mathcal{Y}_s|$-way classifier.
$\mathcal{H}(\cdot)$ measures the entropy of predicted scores of $f_{d}(\boldsymbol{x})$.
Notably, a higher entropy denotes that the input image is more likely from the unseen domain, vice versa.
In this paper, an image is regarded to be from the unseen domain once its entropy is higher than $\tau$.

As only $\mathcal{S}$ is available for training, we target to make $f_{d}(\boldsymbol{x})$ discriminative enough to produce a low entropy for most $\boldsymbol{x}_s$, thus many of the unseen images can be filtered out with relatively high entropy in Eq.~\eqref{eq:inf_all}.
Notably, those correctly filtered unseen images $\boldsymbol{x}_u$ are treated with domain-aware recognition in the searching space $\mathcal{Y}_u$ by Eq.~\eqref{eq:inf_all}.
Under this scheme, DVBE focuses on tackling two main problems: a) how to design $f_{d}(\cdot)$ that is highly discriminative for seen classes; and b) how to design $\{f_{v}(\cdot),g(\cdot)\}$ that can bridge the semantic-visual gap for correct unseen class prediction.
Based on the above two problems, DVBE consists of two sub-modules,~\emph{i.e.,} Adaptive Margin Second-order Embedding and Auto-searched Semantic-Visual Embedding.
Detailed architecture is shown in Figure~\ref{fig:method}.

\subsection{Adaptive Margin Second-order Embedding}
The previous works~\cite{atzmon2019adaptive,vyas2018out} have shown that the out-of-distribution images can be filtered by applying an entropy-based detector once the visual representation is discriminative enough for the in-distribution classes.
Therefore, we target to improve the visual discrimination of $f_{d}(\boldsymbol{x})$ to boost the domain detection in Eq.~\eqref{eq:inf_all}.
To this end, an Adaptive Margin Second-order Embedding (AMSE) is developed by exploring discriminative second-order visual statistics and adaptive margin constraints.

Define the input feature as $\boldsymbol{x}\in R^{W\times H\times C}$, where $W$, $H$, and $C$ are the width, height, and channels, respectively.
For the convenience of bilinear operation, we reshape the feature $\boldsymbol{x}\in R^{W\times H\times C}$ to $\boldsymbol{x}\in R^{N\times C}$, where $N=W\times H$.
Based on the feature $\boldsymbol{x}$, AMSE is proposed based on the well-known bilinear pooling~\cite{lin2015bilinear}:
\begin{eqnarray}
\boldsymbol{x}^{bp} = \boldsymbol{x}\otimes \boldsymbol{x},
\label{eq:bp}
\end{eqnarray}
where $\otimes$ denotes the local pairwise interaction along channel by $\boldsymbol{x}^{bp}=\sum^{N}_{n=1} \boldsymbol{x}_n^{\top}\boldsymbol{x}_n$, and bilinear description $\boldsymbol{x}^{bp}$ need be reshaped to a feature vector for the classifier.
However, besides the informative elements, the noisy elements in $\boldsymbol{x}$ are also magnified by $\otimes$ operator,~\emph{e.g.,} a noisy feature element will be interacted with all other elements.

To reduce the negative effect of noisy features in Eq.~\eqref{eq:bp}, a cross-attentive channel interaction is designed to generate a compact representation $f_d(\boldsymbol{x})$ by:
\begin{eqnarray}
f_d(\boldsymbol{x}) = [f_{att}^{s}(\boldsymbol{x}_2)\odot \boldsymbol{x}_1]\otimes[f_{att}^{c}(\boldsymbol{x}_1)\odot \boldsymbol{x}_2],
\label{eq:cap}
\end{eqnarray}
where $\boldsymbol{x}_1 = f_{rd}^{1}(\boldsymbol{x})$ and $\boldsymbol{x}_2 = f_{rd}^{2}(\boldsymbol{x})$ are two reduction layers to project $\boldsymbol{x}$ into different compressed sub-spaces $R^{N\times D} (D\ll C)$.
$f_{att}^{s}(\cdot)$ and $f_{att}^{c}(\cdot)$ are proposed to generate spatial and channel attention maps, respectively.
$\odot$ is the Hadamard product for element-wise multiplication.
As shown  in Figure~\ref{fig:method}, $f_{att}^{s}(\cdot)$ and $f_{att}^{c}(\cdot)$ are deployed in a cross-attentive way to facilitate attention complementarity.
Compared with Eq.~\eqref{eq:bp}, the cross-attentive channel interaction has two main advantages: a) two different attention functions can eliminate the trivial elements effectively; and b) the cross-attentive manner can improve inputs complementarity.

\begin{figure}[t]
	\begin{center}
		\includegraphics[width=0.95\linewidth]{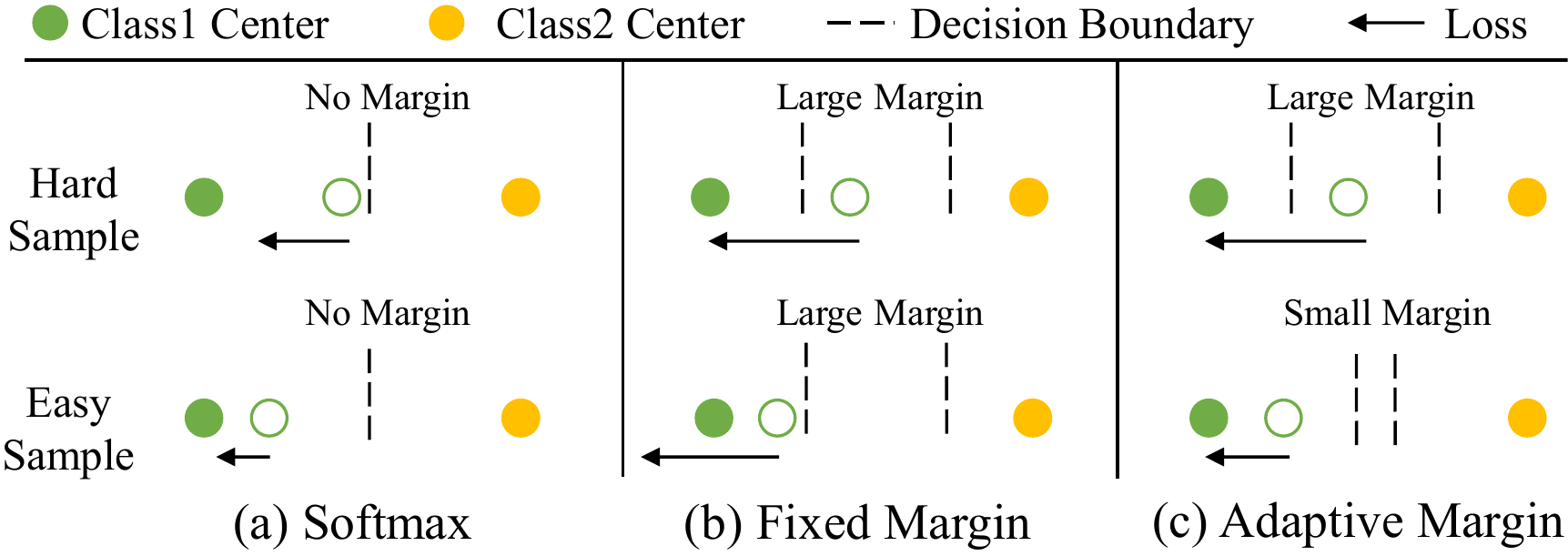}
	\end{center}
	\caption{A toy comparison between different Softmax strategies.}
	\label{fig:toy}
\vspace{-0.3cm}
\end{figure}

To further enlarge the discrimination of $f_{d}(\boldsymbol{x})$,  AMSE then applies an adaptive margin Softmax to maximize the inter-class differences by:
\begin{eqnarray}
\mathcal{L}_{ams} = -\sum_{x\in\mathcal{X}_s} log \frac{e^{\lambda W_yf_d(\boldsymbol{x})}}{e^{\lambda W_yf_d(\boldsymbol{x})}+\sum_{j\in\mathcal{Y}_s,j\neq y}e^{W_jf_d(\boldsymbol{x})}},
\label{eq:L_ams}
\end{eqnarray}
where $W$ is the classifier weight, and $y$ is the groundtruth label. 
$\lambda\in (0,1]$ is a margin coefficient.
When $\lambda=1$, $\mathcal{L}_{ams}$ is the standard Softmax loss.
Differently, $\mathcal{L}_{ams}$ imposes a margin penalty $\lambda$ between $f_{d}(\boldsymbol{x})$ is predicted as $y$ and $\mathcal{Y}\setminus y$.
For example, when $\lambda=0.8$, the response between $\boldsymbol{x}$ and $W_y$ becomes $0.8W_yf_d(\boldsymbol{x})$, which puts a more strict constraint on the inter-class difference to compact samples.

In this work, $\lambda$ is adaptively calculated according to sample difficulties by:
\begin{eqnarray}
\lambda = e^{-(p_y(\boldsymbol{x})-1)^2/\sigma^{2}},
\label{eq:lambda}
\end{eqnarray}
where $p_y(\boldsymbol{x})$ is the predicted probability for $\boldsymbol{x}$ belonging to the class $y$.
Notably, Eq.~\eqref{eq:lambda} exactly follows a Gaussian distribution of mean value of $1$ and variance of $\sigma$.
When $\boldsymbol{x}$ is an easy sample whose $p_y(\boldsymbol{x})\approx 1$, $\mathcal{L}_{ams}$ approximates the standard Softmax loss with $\lambda\approx1$.
Inversely, for the hard samples, $\lambda$ will be small, which indicates a large margin constraint and more attention on this sample.

Compared with fixed margin $\lambda$ in \cite{deng2019arcface}, $\mathcal{L}_{ams}$ adopts a non-linear $\lambda$ in Eq.~\eqref{eq:lambda} to emphasize on those hard samples.
This can prevent a model from being corrupted by a rigorous margin constraint,~\emph{e.g.,} applying a large margin to an easy sample that has been already well-recognized. 
A toy comparison under no margin, fixed margin, and adaptive margin Softmax is given in Figure~\ref{fig:toy}.
Different from the focal loss \cite{lin2017focal}, $\mathcal{L}_{ams}$ utilizes Gaussian margin constraints between decision boundaries, which is more interpretable.
Finally, $\mathcal{L}_{ams}$ enables $f_d(\boldsymbol{x})$ to have a larger inter-class difference than standard Softmax loss.

By exploring cross-attentive second-order statistics with adaptive margin Softmax, the semantic-free $f_d(\boldsymbol{x})$ learned by AMSE has strong visual discrimination for seen classes, which can improve seen class prediction and filter out the unseen images correctly in Eq.~\eqref{eq:inf_all}.

\begin{figure}[t]
	\begin{center}
		\includegraphics[width=0.9\linewidth]{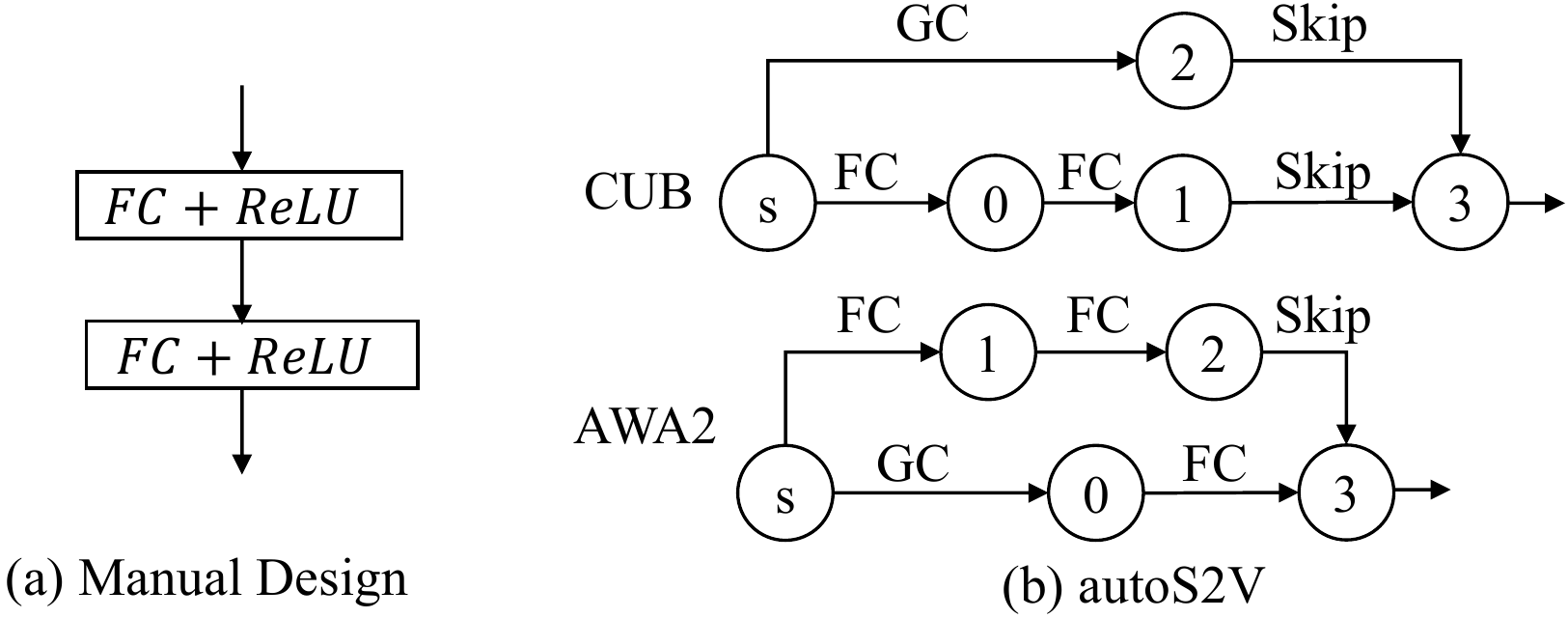}
	\end{center}
	\caption {(a) The hand-designed architecture. b) The learned task-specific architectures from autoS2V for CUB and AWA2, respectively. GC is the graph convolution. }
	\label{fig:autoS2V}
\vspace{-0.3cm}
\end{figure}

\subsection{Auto-searched Semantic-Visual Embedding}
For the unseen image recognition, we develop an auto-searched semantic-visual embedding (AutoS2V) to generate the semantic-aligned visual representation.
Following the general semantic-visual paradigm, AutoS2V consists of $\{f_{v}(\cdot),g(\cdot)\}$ by minimizing $\mathcal{L}_{s2v}$ in Eq.~\eqref{eq:gel_com}.
In most existing methods \cite{Annadani2018,Chen2018}, $g(\cdot)$ is designed manually,~\emph{e.g.,} two FC+ReLU layers that is similar to $f_{v}(\cdot)$, as shown in Figure~\ref{fig:autoS2V} (a).
However, there exists an obvious gap between visual and semantic signals,~\emph{e.g.,} each element in $\boldsymbol{a}$ corresponds to a specific class attribute.
Thus, the processing between $g(\cdot)$ and $f_{v}(\cdot)$ should be different.
In contrast to previous methods, AutoS2V automatically searches an optimal architecture for $g(\cdot)$ with task-specific operations, as shown in Figure~\ref{fig:autoS2V} (b).

The $g(\cdot)$ is viewed as a directed acyclic graph with several nodes, as shown in Figure~\ref{fig:method}.
The input node is omitted for simplification.
Each node is connected with all previous nodes, and operations between two connected nodes are selected from an operation set $\mathcal{O}$.
$\mathcal{O}$ contains four type of operations: (1) fully connection; (2) graph convolution \cite{kampffmeyer2019rethinking,wanglearning,yang2019deep,zhang2014robust}; (3) skip connection; and (4) none.
Notably, graph convolution is a task-specific operation, which is an expert in modeling topology relationships among semantic labels.
To search the optimal operation $\mathcal{O}_{i,j}$ between node $i$ and node $j$, we relax the choice selection as a softmax optimization problem, following \cite{liu2018darts} by:
\begin{eqnarray}
\mathcal{O}_{i,j}=\arg\max_{c\in\mathcal{O}}\frac{exp(\alpha_{i,j}^{c})}{\sum_{o\in\mathcal{O}}exp(\alpha_{i,j}^{o})}, 
\label{eq:nas}
\end{eqnarray}
where $\alpha_{i,j}^c$ is the score for operation $c$.
$\alpha=\{\alpha_{i,j}^c\}$ contains all trainable architecture parameters for $g(\cdot)$.
Compared with hand designs, the automatically searched $g(\cdot)$ is more flexible in bridging the semantic-visual gap, which improves the unseen class prediction.


\subsection{Overall}
Finally, the overall objective function for DVBE is:
\begin{eqnarray}
\mathcal{L}_{all} = \mathcal{L}_{s2v} +  \mathcal{L}_{ams} + \gamma\mathcal{L}_{cet},
\label{eq:L_ovel}
\end{eqnarray}
where $\mathcal{L}_{cet}(f_v(\boldsymbol{x}))$  is an auxiliary cross-entropy loss to avoid all $f_{v}(\boldsymbol{x})$ collapsing into a single point with a control weight $\gamma$.
Besides the object classification task, we further extend the proposed DVBE to the challenging zero-shot semantic segmentation~\cite{NIPS2019_8338}.
Detailed illustrations for segmentation and $\mathcal{L}_{cet}$ are given in supplementary material.

\begin{table}
\begin{center}
\caption{The train and val sets contain seen classes, while the test set contains unseen classes.} \label{tab:data}
\resizebox{1\columnwidth}{!}{
\begin{tabular}{lcccccc}
  \hline
  Datasets &Attributes&$|\mathcal{Y}_s|$&$|\mathcal{Y}_u|$&Train&Val&Test \\
  \hline
  \hline
  CUB~\cite{WelinderEtal2010}&312&150&50&7,057&1,764&2,967\\
  AWA2~\cite{Xian2018}&85&40&10&23,527&5,882&7,913\\
  aPY~\cite{Farhadi2009}&64&20&12&5,932&1,483&7,924\\
  SUN~\cite{Patterson2012}&102&645&72&10,320&2,580&1,440\\
  \hline
\end{tabular}}
\end{center}
\vspace{-0.5cm} 
\end{table}


\section{Experiments}
\begin{table*}
\begin{center}
\caption{Results of GZSL on four classification benchmarks. Generative methods (GEN) utilizes extra synthetic unseen domain data for training. Since many previous methods cannot be end-to-end trained, we define DVBE and DVBE* as fixing and finetuning the backbone weights, respectively. $\dag$ indicates the prediction ensemble from global and local regions.} \label{tab:gzsl}
\resizebox{1.8\columnwidth}{!}{ 
\begin{tabular}{l|l|c|c|c|c|c|c|c|c|c|c|c|c}
  \hline
&\multirow{2}{*}{Methods}&\multicolumn{3}{c|}{CUB~\cite{WelinderEtal2010}}&\multicolumn{3}{c|}{AWA2~\cite{Xian2018}}&\multicolumn{3}{c|}{aPY~\cite{Farhadi2009}}&\multicolumn{3}{c}{SUN~\cite{Patterson2012}}\\
\cline{3-14}
&&\textit{MCA}$_u$&\textit{MCA}$_s$&$H$&\textit{MCA}$_u$&\textit{MCA}$_s$&$H$&\textit{MCA}$_u$&\textit{MCA}$_s$&$H$&\textit{MCA}$_u$&\textit{MCA}$_s$&$H$\\
\hline
\hline
\multirow{3}{*}{\rotatebox{270}{GEN}}&FGN\cite{xian2018feature}&43.7&57.7&49.7&-&-&-&-&-&-&42.6&36.6&39.4\\
&SABR-I\cite{paul2019semantically}&55.0&58.7&56.8&30.3&93.9&46.9&-&-&-&50.7&35.1&41.5\\
&f-VAEGAN-D2\cite{Xian_2019_CVPR}&63.2&75.6&68.9&-&-&-&-&-&-&50.1&37.8&43.1\\
\hline
\hline
\multirow{12}{*}{\rotatebox{270}{NON-GEN}}&CDL\cite{Jiang_2018_ECCV}&23.5&55.2&32.9&-&-&-&19.8&48.6&28.1&21.5&34.7&26.5\\
&PSR-ZSL\cite{Annadani2018}&24.6&54.3&33.9&20.7&73.8&32.2&13.5&51.4&21.4&20.8&37.2&26.7\\
&SP-AEN\cite{Chen2018}&34.7&70.6&46.6&23.3&90.9&37.1&13.7&63.4&22.6&24.9&38.6&30.3\\
&DLFZRL\cite{tong2019hierarchical}&-&-&37.1&-&-&45.1&-&-&31.0&-&-&24.6\\
&MLSE\cite{ding2019marginalized}&22.3&71.6&34.0&23.8&83.2&37.0&12.7&74.3&21.7&20.7&36.4&26.4\\
&TripletLoss\cite{Cacheux_2019_ICCV}&55.8&52.3&53.0&48.5&83.2&61.3&-&-&-&47.9&30.4&36.8\\
&COSMO\cite{atzmon2019adaptive}&44.4&57.8&50.2&-&-&-&-&-&-&44.9&37.7&41.0\\
&PREN*\cite{ye2019progressive}&32.5&55.8&43.1&32.4&88.6&47.4&-&-&-&35.4&27.2&30.8\\
&VSE-S*\cite{zhu2019generalized} &33.4&87.5&48.4&41.6&91.3&57.2&24.5&72.0&36.6&-&-&-\\
&AREN*\dag\cite{xie2019attentive}&63.2&69.0&66.0&54.7&79.1&64.7&30.0&47.9&36.9&40.3&32.3&35.9\\
\cline{2-14}
&DVBE&53.2&60.2&56.5&63.6&70.8&67.0&32.6&58.3&41.8&45.0&37.2&40.7\\
&DVBE*&64.4&73.2&\textbf{68.5}&62.7&77.5&\textbf{69.4}&37.9&55.9&\textbf{45.2}&44.1&41.6&\textbf{42.8}\\
\hline
\end{tabular}}
\end{center}
\vspace{-0.5cm} 
\end{table*}

\subsection{Experimental Settings}
\noindent\textbf{Datasets.}
For generalized zero-shot classification, four widely used benchmarks are evaluated, which are Caltech-USCD Birds-200-2011 (CUB) \cite{WelinderEtal2010}, SUN \cite{Patterson2012}, Animals with Attributes2 (AWA2) \cite{Xian2018}, and Attribute Pascal and Yahoo (aPY) \cite{Farhadi2009}.
All these datasets use the provided category attributes as semantic labels.
The split of seen/unseen classes follows the newly proposed settings in \cite{Xian2018}, which ensures that unseen classes are strictly unknown in the pretrained models.
The details about each dataset are listed in Table~\ref{tab:data}.
For zero-shot semantic segmentation, Pascal VOC \cite{everingham2010pascal} is further evaluated, which follows the setting in \cite{NIPS2019_8338} by splitting the $20$ classes into $14/6$ as seen and unseen classes.
The semantic labels for VOC are generated by using `word2vec' model \cite{mikolov2013distributed}, which is learned on Wikipedia corpus and can produce a $300$ dimensional description.
During training, the images that contain unseen class pixels are all removed.

\noindent\textbf{Implementation Details.}
For classification, the backbone network is ResNet-101 \cite{he2016deep}, which is pre-trained on ImageNet \cite{Russakovsky2015}. 
The rest part of DVBE uses MSRA random initializer.
For data augmentation, random cropping of $448\times 448$ and horizontal flipping are used.
The testing images are cropped from the center part, and averaged horizontal flipping results are reported.
To train the auto-searched $g(\cdot)$, a two-stage training strategy is adopted.
First, the backbone is fixed, and DVBE is trained by alternately updating $\alpha$ and the rest model weights.
Then, $\alpha$ is fixed, and the whole DVBE is finetuned end-to-end.
The batch size is 24, and the reduction channel is $D=256$.
The SGD optimizer is used with initial $lr=0.001$, momentum=0.9, and $180$ training epoch.
The hyper-parameter of $\sigma$ is set to be $0.5$ for most cases, and $\tau$ will be analyzed later.
For semantic segmentation, the backbone network is based on Deeplab v3+ \cite{chen2017rethinking}, as well as the data augmentation and optimizer.
Code is available at \url{github.com/mboboGO/DVBE}.


\noindent\textbf{Metrics.}
Similar to \cite{Xian2018}, the harmonic mean, denoted by  $H = (2MCA_u\times MCA_{s})/(MCA_u+MCA_s)$, is used to evaluate the comprehensive performance of a model in two domains.  $MCA_s$ and $MCA_u$ are the Mean Class Top-1 Accuracy for seen and unseen domains.
For semantic segmentation, similar hIoU is defined following \cite{NIPS2019_8338}.

\subsection{Comparison with state-of-the-art methods}

\noindent\textbf{Object Classification}
The results on CUB, AWA2, aPY, and SUN are given in Table~\ref{tab:gzsl}.
In terms of non-generative models, the proposed DVBE surpasses them by $2.5\%$, $4.7\%$, $8.3\%$, and $1.8\%$ for CUB, AWA2, aPY, and SUN, respectively.
Among the compared methods, AREN obtains the best performance on most of datasets by aggregating the results from global and local regions.
Compared with AREN, DVBE only considers global representation and achieves an averaged 5.6\% improvement on four datasets.
Except for AREN, DVBE outperforms other global-based methods by an averaged $8.5\%$ improvement. 
Notably, the impressive performance of DVBE is mainly attributed to the large improvements in the unseen domain, \emph{e.g.,} 8.0\% and 7.9\% improvements of $MCA_u$ on AWA2 and aPY.
This demonstrates the effectiveness of DVBE in alleviating the biased recognition problem.

Among these methods, COSMO~\cite{atzmon2019adaptive} is the most similar work to DVBE with domain detection mechanism.
From Table~\ref{tab:gzsl}, we can see that the proposed DVBE outperforms COSMO by $6.3\%$ improvement on CUB and comparable performance on SUN.
After end-to-end training, VBDE surpasses COSMO by respective $18.3\%$ and $1.8\%$ improvements on CUB and SUN.
The main reason is that AMSE can significantly improve the visual discrimination of semantic-free representation, thereby resulting in more accurate domain detection than COSMO.
Besides, DVBE adopts a simple entropy-based domain detector, which makes DVBE end-to-end trainable.
Notably, the relatively small improvement on SUN dataset is because that each class contains fewer images than other datasets, which makes it hard to obtain highly discriminative $f_{d}(\boldsymbol{x})$.

As the generative methods employ the prior unseen semantic labels to generate massive synthetic data, they can obtain better performance than the existing non-generative methods. 
In terms of brand-new f-VAEGAN-D2 \cite{Xian_2019_CVPR}, we can observe that DVBE still obtains comparable performance without using unseen domain knowledge.

\begin{figure}
	\begin{center}
		\includegraphics[width=0.95\linewidth]{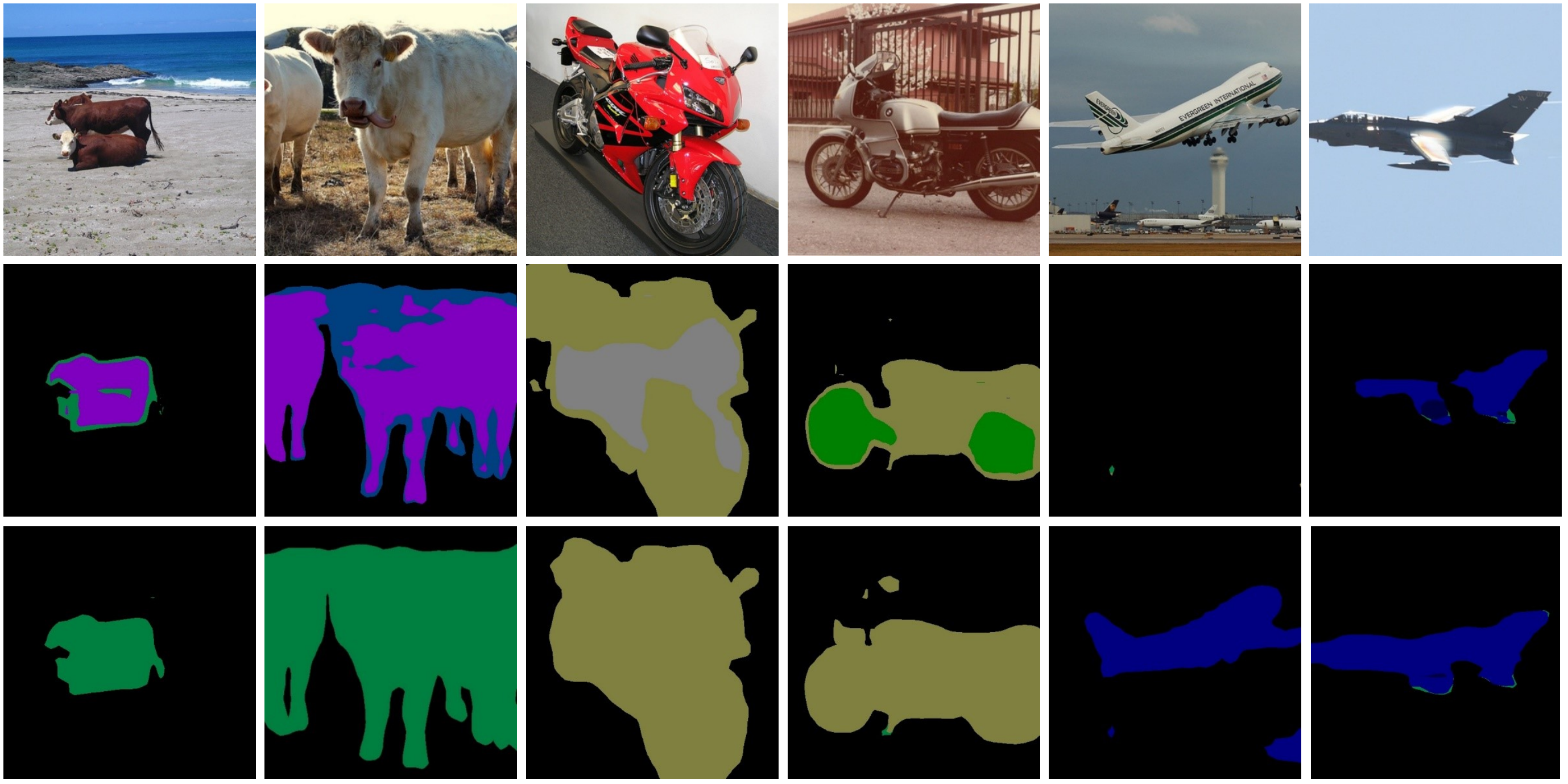}
	\end{center}
	\caption{Some results of zero-shot semantic segmentations. The second and third rows are SP-AEN and DVBE, respectively.}
	\label{fig:seg}
\vspace{-0.3cm}	
\end{figure}

\noindent\textbf{Semantic Segmentation}
We further evaluate the proposed method on semantic segmention task, and summarize the related results in Table~\ref{tab:seg}.
The results of SP-AEN are obtained by extending its public code to ZSL segmentation. 
ZS3Net \cite{NIPS2019_8338} is a newly proposed generative-based method for zero-shot semantic segmentation, and we use their reported performance.
Impressively, DVBE outperforms ZS3Net by a large margin of $11.0\%$ in terms of hIoU.
The reason is that the semantic labels from word2vec contain more noise than manually annotated labels in classification, which makes unseen data synthesis of ZS3Net much harder.
Thus, ZS3Net suffers from serious biased recognition problem,~\emph{e.g.,} a large gap between $26.1\%$ and $69.3\%$ mIoU in unseen and seen domains.
By learning discriminative semantic-free representation, the domain detection by DVBE can effectively alleviate biased recognitions.
Especially for the unseen domain, the mIoU of DVBE is higher than ZS3Net,~\emph{i.e.,} 45.4\% $vs$ 26.1\%.
In addition, we visualize some results of DVBE and SP-AEN in Figure~\ref{fig:seg} to prove the effectiveness of semantic-free representations\footnote{We cannot visualize the results for ZS3Net due to unavailable codes.}.
It can be seen that, with accurate domain detection, the regions of unseen classes can be accurately localized by DVBE, and SP-AEN usually provides biased recognition towards the seen classes. 

\begin{table}
\begin{center}
\caption{GZSL results on Pascal-VOC.}
\label{tab:seg}
\resizebox{0.9\columnwidth}{!}{ 
\begin{tabular}{l|c|c|c|c|c|c}
  \hline
  \multirow{2}{*}{Methods}&\multicolumn{2}{c|}{Seen}&\multicolumn{2}{c|}{Unseen}&\multicolumn{2}{c}{Over} \\
  \cline{2-7}
  &PA&mIoU&PA&mIoU&hPA&hIoU\\
  \hline
  \hline
  Devise\cite{Frome2013}&89.9&64.3&10.3&2.9&18.5&5.5\\
  SP-AEN\cite{Chen2018}&90.5&69.5&15.9&10.5&27.0&18.2\\
  ZS3Net\cite{NIPS2019_8338}&92.9&69.3&46.7&26.1&62.2&37.9\\
  DVBE&89.7&53.1&60.2&45.4&\textbf{72.0}&\textbf{48.9}\\
  \hline
\end{tabular}}
\end{center}
\vspace{-0.5cm} 
\end{table}

\subsection{Ablation Studies}
In this section, we analyze the effects of different components in DVBE.
The BaseS2V is a semantic-visual baseline, which consists of only $\{f_v(\cdot),g(\cdot)\}$ based on Eq~\eqref{eq:gel_inf}, and $g(\cdot)$ is implemented by two layer FC+ReLU.
To evaluate the domain detection performance, we regard GZSL as a binary classification task and define $R_s$ and $R_u$ as the Recall in respective seen and unseen domains.
For example, $R_u$ represents the percentage of how many unseen images are recognized as unseen classes.
Some results are reported in supplementary materials, including attention visualization and hyper-parameters evaluation.

\begin{table}
\begin{center}
\caption{The effects of semantic-free representation. Base $f_{d}(\boldsymbol{x})$ indicates a Conv+ReLU layer. CSE denotes the cross-attentive second-order embedding. $H_R$ is the harmonic mean of $R_s$ and $R_u$. The backbone is fixed, and autoS2V is not adopted here.}
\label{tab:dve}
\resizebox{1\columnwidth}{!}{ 
\begin{tabular}{l||ccc||ccc}
  \hline
  Modules&$MCA_s$&$MCA_u$&$H$&$R_s$&$R_u$&$H_R$ \\
  \hline
  \hline
  BaseS2V&62.5&32.8&43.0&89.5&45.8&60.6\\
  +$f_{d}(\boldsymbol{x})$&51.9&46.6&49.1&61.4&72.8&66.6\\
  +CSE&56.5&50.3&53.2&70.4&78.7&74.3\\
  +$\mathcal{L}_{ams}$(DVBE)&59.9&50.9&\textbf{55.0}&71.6&80.2&\textbf{75.7}\\
  \hline
\end{tabular}}
\end{center}
\vspace{-0.5cm} 
\end{table}

\begin{figure}
    \centering 
    \subfigure[The entropy distributions of seen and unseen images by DVBE.]{ 
    \begin{minipage}[t]{1\linewidth}
    \centering 
    \includegraphics[width=1\linewidth]{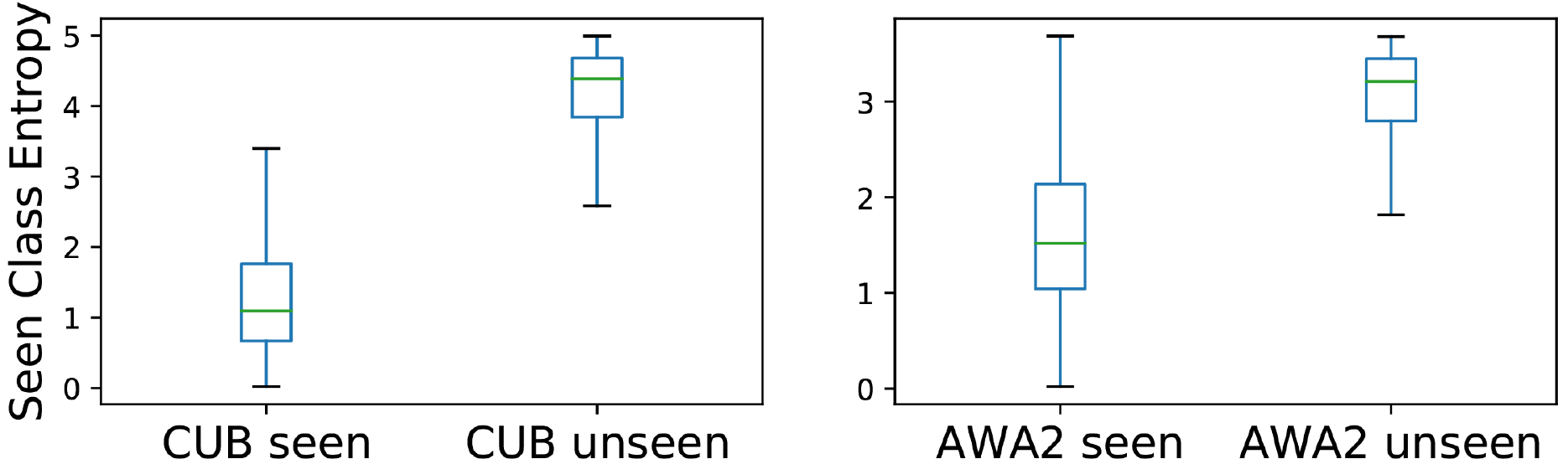} 
    \end{minipage} 
    }

    \subfigure[Evaluation of varying $\tau$ for $MCA_s$, $MCA_u$, and $H$.]{ 
    \begin{minipage}[t]{1\linewidth}
    \centering 
    \includegraphics[width=1\linewidth]{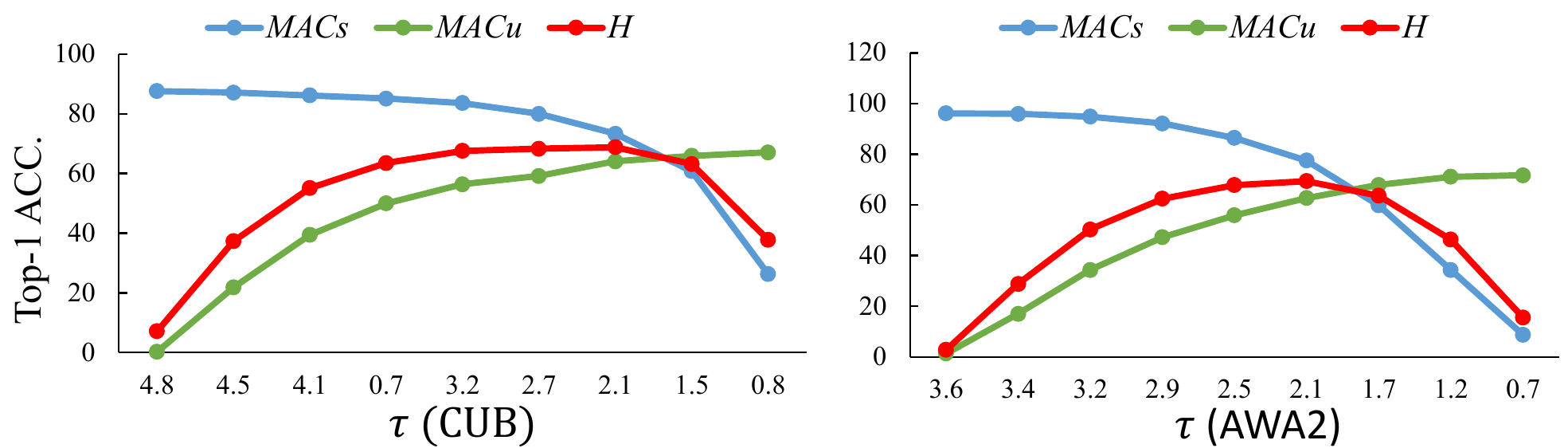} 
    \end{minipage} 
    } 
    \caption{The analysis of domain detection by DVBE.}
    \label{fig:tau}
\vspace{-0.3cm}	
\end{figure}

\noindent\textbf{Effects of $\tau$.}
In this work, $\tau$ is a critical parameter to divide seen and unseen images based on the predicted seen class entropy in Eq.~\eqref{eq:inf_all}. 
We thus calculate the entropy statistics of seen and unseen domain images of DVBE, and give some analyses about $\tau$.
From Figure~\ref{fig:tau} (a), we can see that the seen and unseen images from both CUB and AWA2 have separable entropy statistics.
Thus, the entropy-based detector can well separate them apart using an appropriate threshold of $\tau$.
Usually, a higher $\tau$ means that DVBE may regard more unseen images as the seen domain, vice versa.
Thus, different $\tau$ has different effects on domain detection.
As shown in Figure~\ref{fig:tau} (b), when increasing $\tau$, $MCA_{s}$ and $MCA_{u}$ have opposite changing tendencies.
By taking $H$ into consideration, we set $\tau$ to achieve a trade-off between $MCA_{u}$ and $MCA_{s}$.
Since different datasets have different category numbers and difficulties, $\tau$ is set to be $2.1$, $2.1$, $1.5$, and $3.9$ for respective CUB, AWA2, aPY, and SUN.

  

\begin{figure}
	\begin{center}
		\includegraphics[width=0.95\linewidth]{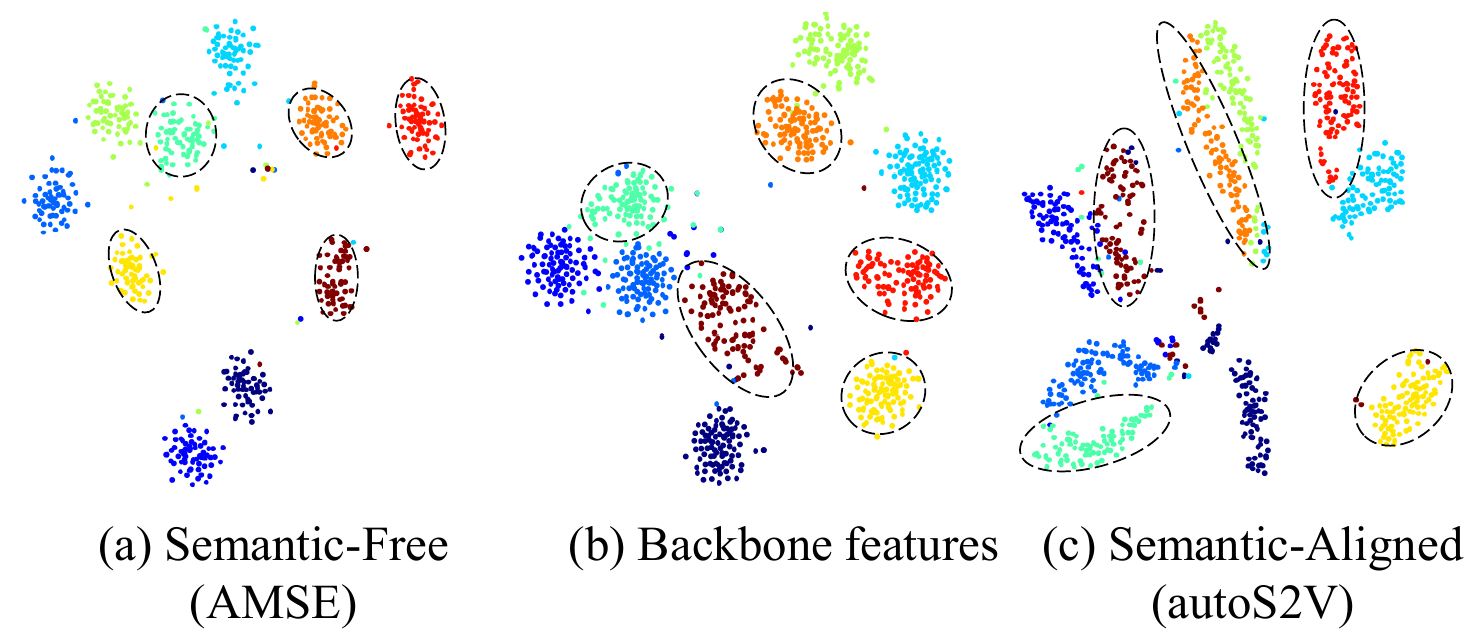}
	\end{center}
	\caption{Representation distributions with/without semantics on AWA2. The dashed circles indicate seen classes. }
	\label{fig:tsne}
\vspace{-0.3cm}	
\end{figure}

\noindent\textbf{Effects of semantic-free visual representation.}
We then explore the effects of semantic-free visual representation in GZSL.
Compared with BaseS2V, DVBE learns complementary semantic-free visual representation $f_{d}(\boldsymbol{x})$ to predict seen classes and filter out the unseen images.
As shown in Table~\ref{tab:dve}, by adding the semantic-free $f_{d}(\boldsymbol{x})$ to BaseS2V, $H_R$ is improved from $60.6\%$ to $66.6\%$. 
Notably, $H_R$ measures the domain detection performance of a GZSL model, thus the improved $H_R$ proves that the semantic-free $f_d(\boldsymbol{x})$ can effectively alleviate the biased recognition problem, and $H$ is improved from $43.0\%$ to $49.1\%$.
When further improving the visual discrimination of $f_d(\boldsymbol{x})$ via CSE and $\mathcal{L}_{ams}$, both $H_R$ and $H$ are increasing. 
Besides, we also visualize the feature distributions before and after applying AMSE to $f_{d}(\cdot)$ in Figure~\ref{fig:tsne} (a) and (b).
It can be seen that AMSE can significantly improve the visual discrimination of $\boldsymbol{x}$.
Consequently, DVBE outperforms BaseS2V by $12.0\%$ gain.

\noindent\textbf{Cross-attentive channel interaction.}
We next evaluate the effects of different components in cross-attentive channel interaction.
As shown in Table~\ref{tab:cap}, using the bilinear pooling operation $\otimes$ obtains $1.7\%$ improvement of $H$ upon the base $f_d(\boldsymbol{x})$.
Further, by using $f_{att}^{s}(\cdot)$ and $f_{att}^{c}(\cdot)$ to generate attentive features, $H$ is improved from $50.8\%$ to $52.4\%$, which shows that the redundancy reduction can boost the visual discrimination of $f_d(\boldsymbol{x})$.
Finally, the cross-attentive manner also brings improvement by enhancing the inputs complementarity for second-order representation.

\begin{table}
\begin{center}
\caption{Analysis of cross-attentive channel interaction on CUB.}
\resizebox{0.95\columnwidth}{!}{ 
\label{tab:cap}
\begin{tabular}{l||cccc||ccc}
  \hline
  &$\bigotimes$&$f_{att}^{s}(\cdot)$&$f_{att}^{c}(\cdot)$&cross&$MCA_s$&$MCA_u$&$H$ \\
  \hline
  \hline
  \multirow{4}*{$f_d(\boldsymbol{x})$}&&&&&51.9&46.6&49.1\\
  &$\surd$&&&&53.8&48.1&50.8\\
  &$\surd$&$\surd$&$\surd$&&55.1&50.0&52.4\\
  &$\surd$&$\surd$&$\surd$&$\surd$&56.5&50.3&$\textbf{53.2}$\\
  \hline
\end{tabular}}
\end{center}
\vspace{-0.5cm} 
\end{table}

\begin{figure}[t]
	\begin{center}
		\includegraphics[width=0.9\linewidth]{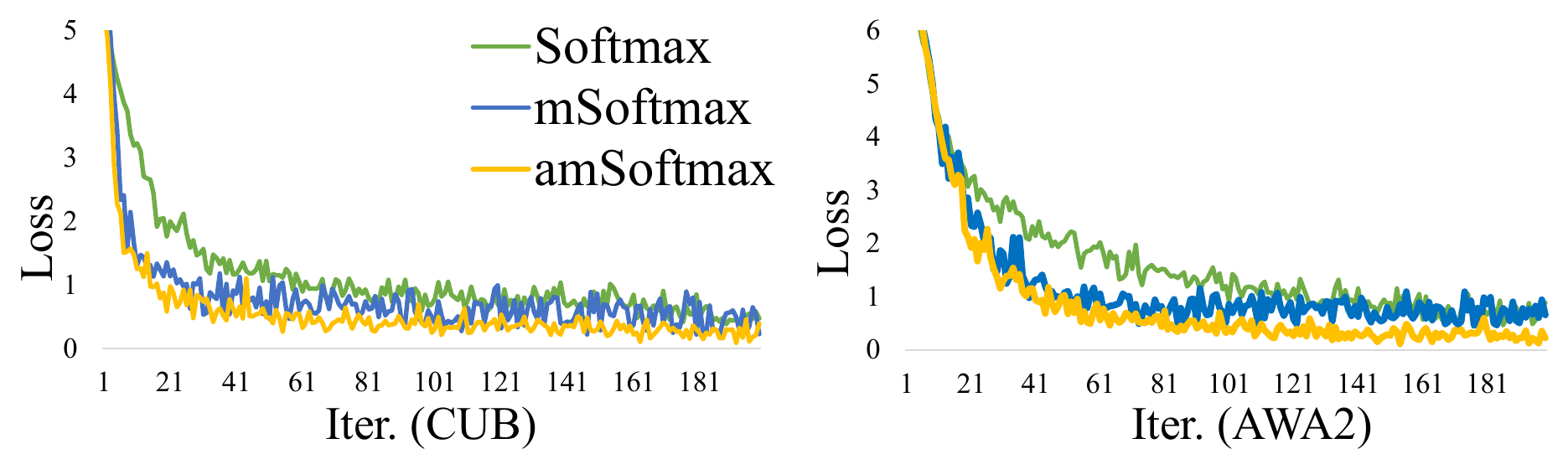}
	\end{center}
	\caption{The training curves of different Softmax strategies.}
	\label{fig:curve}
\vspace{-0.1cm}	
\end{figure}

\noindent\textbf{Adaptive margin Softmax.}
We further analyze the effect of adaptive margin Softmax $\mathcal{L}_{ams}$. 
As shown in Table~\ref{tab:softmax}, the adaptive margin Softmax obtains a higher performance than standard and fixed margin Softmax.
Specifically, imposing a fixed margin on Softmax obtains a lower performance than standard Softmax.
Different from using a fixed margin, we adaptively adjust margin $\lambda$ and achieve the best performance.
In Figure~\ref{fig:curve}, the training curves for different Softmax strategies show that $\mathcal{L}_{ams}$ obtains both fast and stable convergence.



\begin{figure}
	\begin{center}
		\includegraphics[width=0.9\linewidth]{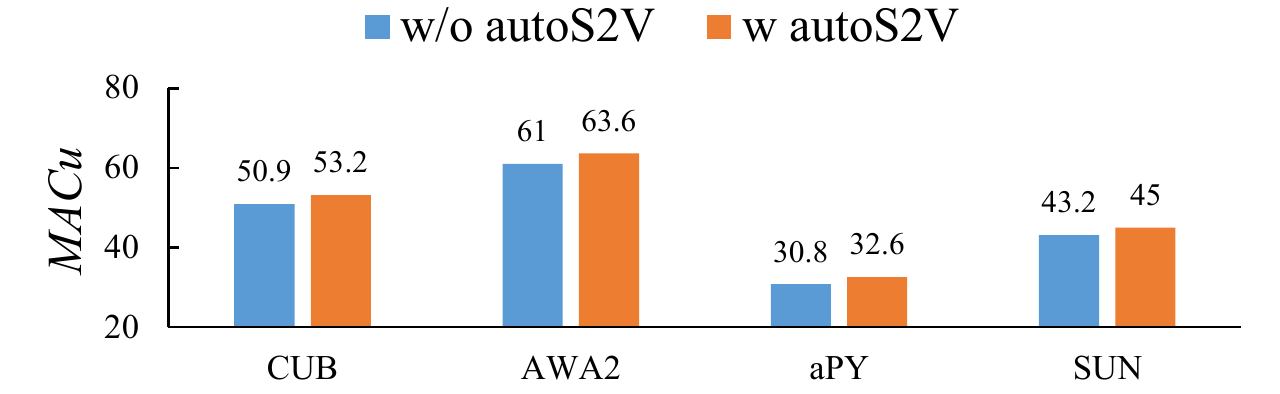}
	\end{center}
	\caption{The improvements obtained by autoS2V on four datasets.}
	\label{fig:autoR}
\vspace{-0.3cm}	
\end{figure}

\noindent\textbf{Auto-searched $g(\cdot)$.}
We finally demonstrate that using autoS2V can generate the optimal and task-related architectures $g(\cdot)$.
As shown in Figure~\ref{fig:autoR}, the auto-searched $g(\cdot)$ brings further improvements in terms of $MCA_u$ than hand-designed one, and the auto-searched $g(\cdot)$ for CUB and AWA2 have different structures in Figure~\ref{fig:autoS2V}.
By further analyzing the obtained architectures, we have the following observations: a) the graph convolution, which is used to explore semantic topology, tends to be used once due to the over-smoothing problem \cite{kampffmeyer2019rethinking}; and b) two branch embeddings are preferred, which is similar to the existing work~\cite{tong2019hierarchical}.
After semantic alignment, the visual features from two domains are associated more tightly for knowledge transfer, as shown in Figure~\ref{fig:tsne} (b) and (c).

\begin{table}
\begin{center}
\caption{Evaluations of different Softmax results on CUB.}
\resizebox{0.85\columnwidth}{!}{ 
\label{tab:softmax}
\begin{tabular}{l||ccc}
  \hline
  Methods&$MCA_s$&$MCA_u$&$H$ \\
  \hline
  \hline
  Standard Softmax&56.5&50.3&53.2\\
  Fixed Margin Softmax&58.2&47.1&52.1\\
  Adaptive Margin Softmax&59.9&50.9&\textbf{55.0}\\
  \hline
\end{tabular}}
\end{center}
\vspace{-0.5cm}	
\end{table}

\section{Conclusion}
In this paper, we propose a novel Domain-aware Visual Bias Eliminating (DVBE) network to solve the biased recognition problem in generalized zero-shot learning (GZSL).
Different from previous methods that focus on semantic-aligned representations, we consider the effect of semantic-free representation in alleviating the biased recognition by detecting unseen images based on seen class prediction entropy.  
To further boost the visual discrimination of semantic-free representations, we develop an adaptive margin second-order embedding via cross-attentive channel interaction and adaptive margin Softmax.
Besides, we automatically search the optimal semantic-visual architectures to produce robust semantic-aligned representation.
Consequently, by exploring complementary visual representations,~\emph{i.e.,} semantic-free and semantic-aligned, DVBE outperforms existing methods by a large margin in terms of both classification and segmentation.

In the future, we target to explore elaborate domain detector to improve DVBE.  

\section*{Acknowledgments}
This work is supported by the National Key Research and Development Program of China (2017YFC0820600), the National Nature Science Foundation of China (61525206, U1936210, 61902399), the Youth Innovation Promotion Association Chinese Academy of Sciences (2017209), and the Fundamental Research Funds for the Central Universities under Grant WK2100100030.

{\small
\bibliographystyle{ieee_fullname}
\bibliography{ref}
}

\end{document}